\documentclass[conference, a4paper]{IEEEtran}
\IEEEoverridecommandlockouts
\IEEEpubid{\makebox[\columnwidth]{\parbox{\columnwidth}{\ \\[12pt]\footnotesize \copyright~2026 IEEE. Personal use of this material is permitted. Permission from IEEE must be obtained for all other uses, in any current or future media, including reprinting/republishing this material for advertising or promotional purposes, creating new collective works, for resale or redistribution to servers or lists, or reuse of any copyrighted component of this work in other works.}}\hspace{\columnsep}\makebox[\columnwidth]{ }}
\usepackage{cite}
\usepackage{amsmath,amssymb,amsfonts}
\usepackage{algorithmic}
\usepackage{graphicx}
\usepackage{textcomp}
\usepackage{booktabs}
\usepackage{multirow}
\usepackage{array}
\usepackage[hidelinks]{hyperref}
\usepackage[table,xcdraw]{xcolor}
\usepackage[
  font=footnotesize,
  justification=centering,
  singlelinecheck=false
]{caption}

\usepackage{titlesec}
\usepackage{geometry}

\usepackage{indentfirst}

\titlespacing*{\section}{0pt}{8pt}{4pt}
\titlespacing*{\subsection}{0pt}{6pt}{3pt}
\titlespacing*{\subsubsection}{0pt}{4pt}{2pt}

\def\BibTeX{{\rm B\kern-.05em{\sc i\kern-.025em b}\kern-.08em
    T\kern-.1667em\lower.7ex\hbox{E}\kern-.125emX}}

\begin{document}

\title{MixedPEFT: Combining Multiple PEFT Methods with Mixed Objectives for Unsupervised Domain Adaptation}

\makeatletter
\def\@IEEEauthorblockNfont{\fontsize{11pt}{13pt}\selectfont}
\def\@IEEEauthorblockAfont{\fontsize{10pt}{12pt}\selectfont}
\makeatother
\author{
\parbox{\textwidth}{
\centering

\begin{tabular}{ccc}
\begin{minipage}[t]{0.29\textwidth}
\centering
{\fontsize{11pt}{13pt}\selectfont Mohammed Rawhani\par}
{\fontsize{10pt}{12pt}\selectfont
\textit{Erciyes University}\par
\textit{Computer Engineering Department}\par
Kayseri, Türkiye\par
mrawhani5@gmail.com}
\end{minipage}
&
\begin{minipage}[t]{0.29\textwidth}
\centering
{\fontsize{11pt}{13pt}\selectfont Dervi\c{s} Karabo\u{g}a\par}
{\fontsize{10pt}{12pt}\selectfont
\textit{Erciyes University}\par
\textit{Computer Engineering Department}\par
Kayseri, Türkiye\par
karaboga@erciyes.edu.tr}
\end{minipage}
&
\begin{minipage}[t]{0.29\textwidth}
\centering
{\fontsize{11pt}{13pt}\selectfont \"{O}zkan Ufuk Nalbanto\u{g}lu\par}
{\fontsize{10pt}{12pt}\selectfont
\textit{Erciyes University}\par
\textit{Computer Engineering Department}\par
Kayseri, Türkiye\par
nalbantoglu@erciyes.edu.tr}
\end{minipage}
\end{tabular}

\vspace{1em}

\begin{tabular}{cc}
\begin{minipage}[t]{0.29\textwidth}
\centering
{\fontsize{11pt}{13pt}\selectfont Alper Ba\c{s}t\"{u}rk\par}
{\fontsize{10pt}{12pt}\selectfont
\textit{Erciyes University}\par
\textit{Computer Engineering Department}\par
Kayseri, Türkiye\par
ab@erciyes.edu.tr}
\end{minipage}
\hspace{0.06\textwidth}
&
\begin{minipage}[t]{0.29\textwidth}
\centering
{\fontsize{11pt}{13pt}\selectfont Bahriye Akay\par}
{\fontsize{10pt}{12pt}\selectfont
\textit{Erciyes University}\par
\textit{Computer Engineering Department}\par
Kayseri, Türkiye\par
bahriye@erciyes.edu.tr}
\end{minipage}
\end{tabular}

}
}
\maketitle
\IEEEpubidadjcol

\begin{abstract}
Applying pre-trained language models to new domains through full fine-tuning is computationally expensive and prone to catastrophic forgetting. To address this limitation, we introduce a novel parameter-efficient strategy for unsupervised domain adaptation that combines a custom PEFT architecture with mixed-objective training. The proposed method integrates invertible adapters with Low-Rank Adaptation (LoRA) and jointly optimizes classification on labeled source-domain data and masked language modeling on unlabeled target-domain data. This joint training scheme supports task adaptation while preserving knowledge of the target domain. We evaluate the method on the Multi-Genre Natural Language Inference (MNLI) dataset across 20 domain shifts. Our approach achieves average performance improvements of 1.41 percentage points over the parameter-efficient state-of-the-art UDapter, 1.26 percentage points over the fully tuned DANN baseline, and 0.86 percentage points over DSN, while updating only 7\% of the model parameters. These findings establish a new state-of-the-art result for parameter-efficient unsupervised domain adaptation and demonstrate that carefully designed PEFT combinations with concurrent optimization can outperform both parameter-efficient and conventional fully tuned approaches.
\end{abstract}


\section{Introduction}
Pre-trained language models (PrLMs), including BERT \cite{devlin2019bert} and RoBERTa \cite{liu2019roberta}, have substantially advanced natural language processing. However, when deployed in real-world settings, these models often encounter domain shifts that can substantially reduce their performance \cite{jiang2007instance}. Unsupervised Domain Adaptation (UDA) addresses this challenge by adapting models to new domains using labeled source data and unlabeled target data \cite{ben2010theory}. Existing UDA methods are generally divided into model-centric approaches, which modify architectures or loss functions \cite{blitzer2006domain,ganin2016domain}, and data-centric approaches that use techniques such as adaptive pretraining  \cite{gururangan2020don,xu2019bert}.

Most existing UDA methods rely on full fine-tuning, which becomes costly as models exceed 100 million parameters \cite{brown2020language}, may cause catastrophic forgetting \cite{kirkpatrick2017overcoming}, and is difficult to scale. Parameter-efficient fine-tuning (PEFT) instead updates only a small subset of parameters while retaining competitive performance \cite{houlsby2019parameter,hu2022lora}. Adapters \cite{pfeiffer2020adapterhub}, Low-Rank Adaptation (LoRA) \cite{hu2022lora}, and prefix tuning \cite{li2021prefix} have shown remarkable success on downstream tasks, but their use in UDA remains limited. Existing studies mainly evaluate individual PEFT methods rather than the potential benefits of combining multiple techniques \cite{malik2023udapter,uppaal2024feuda,zhang2021domain}.

Although PEFT combinations have recently been explored for general fine-tuning \cite{mao2022unipelt}, UDA often requires specialized approaches. Our previous work \cite{rawhani2024efficient} combined invertible adapters with LoRA in a sequential two-phase training strategy for UDA and achieved competitive results among parameter-efficient methods. However, it did not outperform conventional fully tuned baselines and may not fully exploit joint optimization of domain adaptation and task performance, which earlier studies found effective \cite{chronopoulou2019adapters,karouzos2021udalm}.

\begin{figure}[htbp]
\centering
\includegraphics[width=\columnwidth]{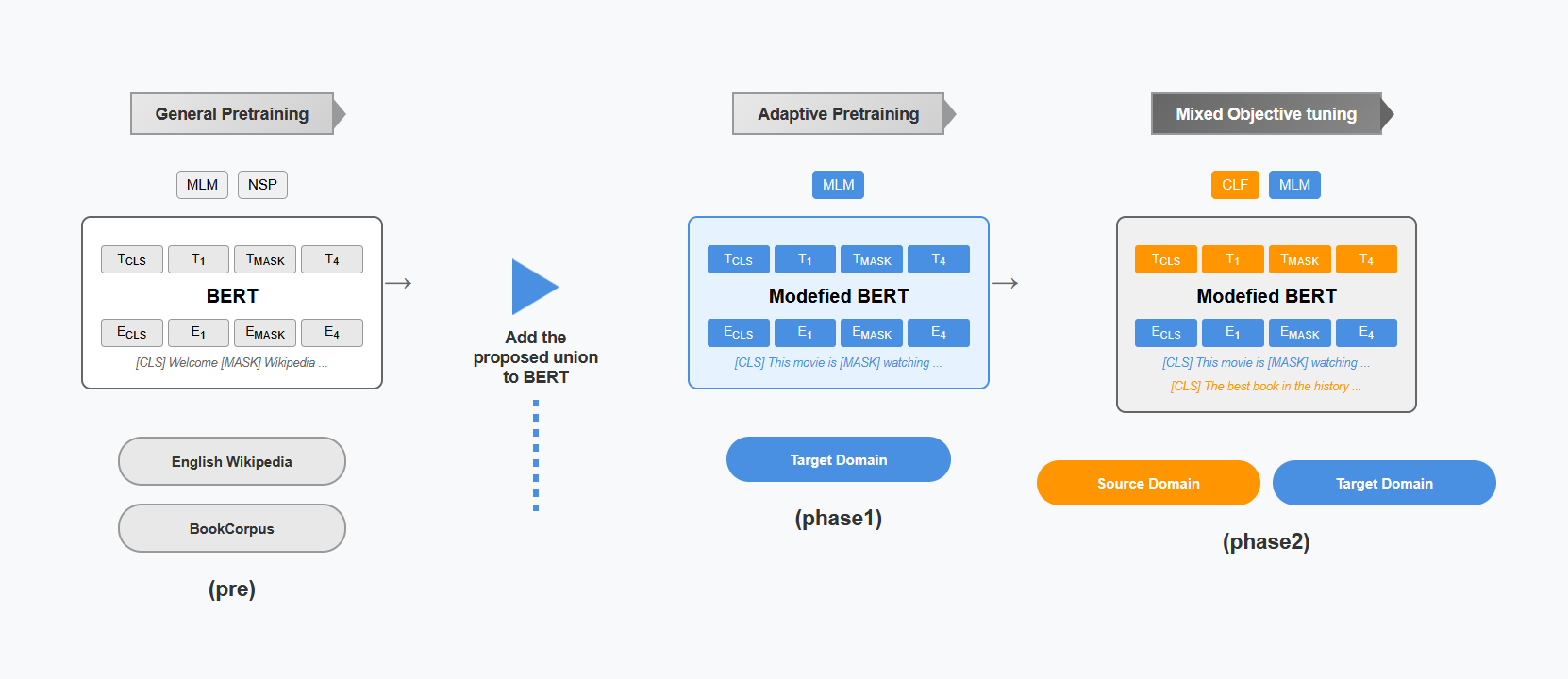}
\caption{Our mixed-objective training process: Phase 1 involves domain-adaptive pretraining with MLM on target-domain data. Phase 2 simultaneously optimizes classification on source-domain data and MLM on target-domain data.}
\label{fig:mixed_objective}
\end{figure}

This work presents a novel parameter-efficient UDA method that combines multiple PEFT components with mixed-objective training. Following the principle of UDALM \cite{karouzos2021udalm}, the model jointly optimizes source-domain classification and target-domain masked language modeling within a parameter-efficient framework with carefully designed PEFT combinations. The overall process is illustrated in Fig.~\ref{fig:mixed_objective}.

Our contributions are threefold: (1) We propose a novel approach specifically designed for parameter-efficient unsupervised domain adaptation. It combines a custom-designed PEFT architecture with mixed-objective training to jointly optimize task performance and domain adaptation objectives. (2) We provide comprehensive experimental validation on the MNLI dataset across 20 domain shifts with different PEFT combinations, showing improvements over existing parameter-efficient and fully tuned baselines while updating only 7\% of the model parameters. (3) We evaluate the methodology with multiple PEFT unification frameworks, supporting its applicability across different PEFT combinations.

\section{Related Work}

\subsection{Unsupervised Domain Adaptation (UDA)}

UDA has been widely studied in the context of deep learning and natural language processing. Traditional approaches can be broadly grouped into model-centric and data-centric methods \cite{ramponi2020neural}. Model-centric methods focus on learning domain-invariant representations through techniques such as adversarial training \cite{ganin2015unsupervised}, domain separation networks \cite{bousmalis2016domain}, and gradient reversal layers \cite{ganin2016domain}. Domain-Adversarial Neural Networks (DANN) \cite{ganin2015unsupervised} learn domain-invariant features by minimizing task loss while maximizing domain confusion through gradient reversal layers. Domain Separation Networks (DSN) \cite{bousmalis2016domain} enhance DANN by incorporating additional losses to preserve domain-specific information while extracting domain-invariant features.

Data-centric methods focus instead on using unlabeled target domain data through techniques such as adaptive pretraining \cite{gururangan2020don}, pseudo-labeling \cite{zou2018unsupervised}, and data augmentation \cite{zhang2015character}. In particular, adaptive pretraining has shown promising results by continuing pretraining on target-domain data before downstream fine-tuning \cite{rietzler2020adapt}. However, most of these methods still require full parameter updates, which limits their use with large-scale models.

Recent work has explored the combination of both approaches. UDALM \cite{karouzos2021udalm} begins with adaptive pretraining with MLM using unlabeled target domain data, then in the second phase, it adds an auxiliary MLM task for target data, jointly with a classification task on labeled source domain, demonstrating the effectiveness of mixed-objective training. This approach maintains domain adaptation capabilities throughout the fine-tuning process, leading to improved performance across various domain adaptation benchmarks.

\subsection{Parameter-Efficient Fine-Tuning}

Parameter-efficient fine-tuning has emerged as a useful technique for adapting pre-trained language models to downstream tasks while minimizing computational costs. Adapter-based methods \cite{pfeiffer2020adapterhub} insert small neural network modules between transformer layers, allowing for task-specific adaptation while keeping the original model parameters frozen. Variants of adapters include bottleneck adapters \cite{pfeiffer2021adapterfusion} and invertible adapters \cite{pfeiffer2020mad}.

Low-Rank Adaptation (LoRA) \cite{hu2022lora} takes a different approach by decomposing weight updates into low-rank matrices, significantly reducing the number of trainable parameters while maintaining competitive performance. 

Recent work has begun to explore combinations of PEFT methods. UniPELT \cite{mao2022unipelt} proposed a unified framework that combines adapters, prefix tuning, and LoRA through a gating mechanism, demonstrating that different PEFT methods can complement each other effectively. MAM Adapter \cite{he2022mam} combined multiple adaptation mechanisms within a single framework, showing that different PEFT methods can complement each other when properly integrated. However, the application of PEFT combinations to unsupervised domain adaptation remains largely unexplored.

\subsection{PEFT for Domain Adaptation}

The intersection of parameter-efficient fine-tuning and unsupervised domain adaptation represents a relatively new research area with significant potential. Early work in this direction includes UDapter \cite{malik2023udapter}, which applied adapter-based methods to domain adaptation tasks, achieving competitive results with reduced parameter counts. Zhang et al. \cite{zhang2021domain} propose an adapter-based unsupervised domain adaptation method that introduces a domain-fusion training step. In a more recent development, Uppaal et al. \cite{uppaal2024feuda} present FEUDA, a minimalist approach to UDA using prompt tuning. However, UDapter and other approaches focused on single adapter types without exploring the potential benefits of combining PEFT methods in domain adaptation, used complex integration with UDA approaches, and remained limited in outperforming traditional fully-tuned methods.

Our previous work \cite{rawhani2024efficient} addressed this gap by proposing a custom union of invertible adapters and LoRA for UDA, using a sequential two-phase training approach. While this method achieved competitive results with parameter-efficient state-of-the-art approaches, the results were only nearly comparable to traditional fully tuned approaches like DANN and DSN. Moreover, the sequential nature of the training process, while simple and straightforward, may not fully exploit the potential for simultaneous optimization of domain adaptation and task performance objectives, which has proven effective in earlier domain adaptation studies.

We address this limitation by proposing an innovative integration between the custom union we developed \cite{rawhani2024efficient} and a mixed-objective training methodology.

\section{Methodology}

\subsection{Overview}

Our approach builds upon the foundation of parameter-efficient fine-tuning by introducing a PEFT-combined mixed-objective training methodology specifically designed for UDA. The key innovation lies in using a combination of complementary PEFT methods to simultaneously optimize two complementary objectives: classification performance on labeled source domain data and masked language modeling on unlabeled target domain data. This approach is implemented through a custom union of parameter-efficient methods that leverages the complementary strengths of invertible adapters and Low-Rank Adaptation (LoRA).

The training process consists of two phases: (1) domain-adaptive pretraining using masked language modeling on target domain data, and (2) mixed-objective fine-tuning that simultaneously optimizes classification and MLM objectives. This design ensures that the model maintains domain-specific knowledge while learning task-relevant features, addressing the key challenges of catastrophic forgetting and parameter efficiency in domain adaptation. 

\subsection{Unification Architecture}

Our previous work \cite{rawhani2024efficient} demonstrated that combining PEFT methods is more effective than using single methods. Moreover, since PEFT approaches involve updating only a small subset of parameters, unifying multiple methods does not significantly increase the overall parameter count, thereby avoiding parameter overload \cite{mao2022unipelt}. Furthermore, different PEFT techniques operate at distinct locations within the transformer architecture, such as attention projections, feed-forward layers, or embedding spaces, making it feasible to combine them without causing architectural interference \cite{mao2022unipelt,he2022mam}.

\subsubsection{Custom Union Configuration}

Our custom union combines invertible adapters with LoRA to create a synergistic parameter-efficient adaptation framework. This combination is motivated by extensive experiments and the complementary nature of the two methods: invertible adapters excel at preserving information while learning domain-specific transformations, while LoRA efficiently captures task-specific adaptations in attention mechanisms, adds minimal computational load which prevents overfitting, and is more effective in NLI tasks than other methods \cite{rawhani2024efficient}:

\begin{equation}
\mathbf{h} = \text{InvertibleAdapter}(\mathbf{x}) + \text{LoRA}(\mathbf{x})
\end{equation}

\begin{figure}[htbp]
\centering
\resizebox{\columnwidth}{!}{%
    \rotatebox{90}{%
        \includegraphics{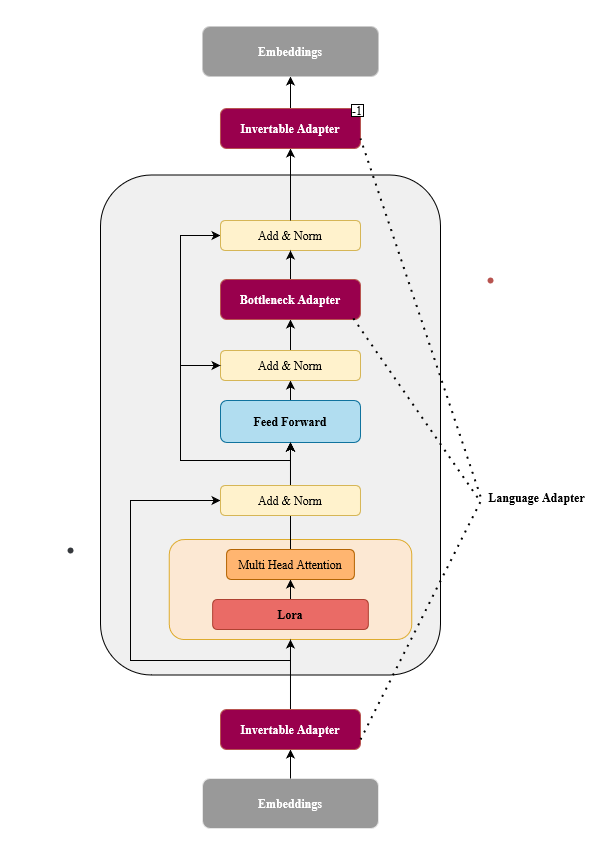}%
    }%
}
\caption{Illustration of our proposed custom union from \cite{rawhani2024efficient}: We combine the LoRA adapter with the invertible adapter, which itself unifies the bottleneck and invertible layers.}
\label{fig:custom_union}
\end{figure}

This combination allows both methods to contribute independently to the final representation while maintaining the residual connection structure of the original transformer. The total parameter count for the custom union is approximately 7.68M parameters, representing only 7.055\% of the BERT-base model's total parameters. Fig.~\ref{fig:custom_union} further illustrates the architecture of our custom union, and more details can be found in \cite{rawhani2024efficient}.

\subsection{Mixed-Objective Training}

As shown in Fig.~\ref{fig:mixed_objective}, our approach generally consists of two phases: 

\subsubsection{Phase 1: Domain-Adaptive Pretraining}

The first phase performs continued pretraining on unlabeled target domain data using the masked language modeling objective. This phase adapts the model's representations to target domain characteristics while maintaining linguistic knowledge. Only the PEFT parameters are updated while base model parameters remain frozen.

The MLM objective minimizes the negative log-likelihood:

\begin{equation}
\mathcal{L}_{\text{MLM}} = -\sum_{i \in \mathcal{M}} \log P(x_i | \mathbf{x}_{\text{masked}})
\end{equation}

where $\mathcal{M}$ denotes the set of masked token positions and $\mathbf{x}_{\text{masked}}$ represents the corrupted input sequence.

\subsubsection{Phase 2: Mixed-Objective Fine-tuning}

The second phase implements the key innovation of our approach: mixed-objective training that simultaneously optimizes classification and masked language modeling during fine-tuning while properly integrating with PEFT combinations. The combined loss function is defined as:

\begin{equation}
\mathcal{L}_{\text{total}} = \alpha \times \mathcal{L}_{\text{classification}} + (1-\alpha) \times \mathcal{L}_{\text{MLM}}
\end{equation}

where $\mathcal{L}_{\text{classification}}$ is the cross-entropy loss for the classification task:

\begin{equation}
\mathcal{L}_{\text{classification}} = -\sum_{i=1}^{N} \sum_{j=1}^{C} y_{ij} \log(\hat{y}_{ij})
\end{equation}

The weighting parameter $\alpha$ is dynamically adjusted based on the relative sizes of source and target datasets:

\begin{equation}
\alpha = \frac{n_{\text{source}}}{n_{\text{source}} + n_{\text{target}}}
\end{equation}

This dynamic weighting ensures balanced contribution from both objectives based on data availability, preventing one objective from dominating the training process. Fig.~\ref{fig:mixed_objective} illustrates our mixed-objective training process.

\section{Experimental Setup}

\subsection{Dataset and Evaluation Protocol}

We conduct comprehensive experiments on the Multi-Genre Natural Language Inference (MNLI) dataset \cite{williams2018broad}, which provides a robust benchmark for evaluating UDA methods. MNLI contains premise-hypothesis pairs from multiple genres, making it ideal for studying domain shift effects in natural language inference tasks.

Our evaluation protocol follows established domain adaptation practices and covers 20 domain shifts. Each domain shift represents adaptation from one genre to another, providing diverse scenarios that test the robustness and generalizability of our approach.

\subsection{Baseline Methods}

We compare our approach against several categories of baseline methods:

\textbf{Parameter-Efficient Baselines:} We include UDApter \cite{malik2023udapter}, which represents the current state of the art in parameter-efficient UDA, and our previous approach \cite{rawhani2024efficient}, which achieved a higher average score than UDApter.

\textbf{Fully-Tuned Traditional Methods:} DANN~\cite{ganin2015unsupervised} and DSN~\cite{bousmalis2016domain} are well-established UDA methods that require full fine-tuning of 100\% of the model parameters. Additionally, we include the \textbf{upper bound (UB)}, which represents fully supervised training on the complete labeled target dataset.

\subsection{Implementation Details}

The implementation utilizes BERT-base-uncased as the foundation model with 109,514,298 total parameters. The custom union configuration combines invertible adapter with reduction factor 2 and LoRA with rank 8 and alpha 16, resulting in 7,682,688 total trainable parameters (7.055\% of base model). Training hyperparameters include learning rate 1e-4 with AdamW optimizer, batch size 32 with gradient accumulation steps 4, and 10 training epochs for the mixed-objective phase.

To ensure comparability, our implementation follows the experimental protocol of UDApter~\cite{malik2023udapter} and our previous work~\cite{rawhani2024efficient}, including the MNLI domain-shift setup, macro-F1 evaluation metric, random seed configuration, and hyperparameter settings. The baseline scores for UDApter, DANN, and DSN are taken from the UDApter study, which reports these methods under the same evaluation protocol. For our proposed variants, all deviations from the UDApter setup are explicitly reported.

\section{Results and Discussion}

We evaluate the mixed-objective approach using two PEFT unification frameworks: our custom union and UniPELT. This comparison examines whether the proposed integration remains effective across different mixed-PEFT architectures. We compare our two variants against all baselines.

\subsection{Comparison with UDApter and Previous Approach}
Table~\ref{tab:udapter_comparison} compares the two variants with UDApter \cite{malik2023udapter} and our previous sequential custom-union approach \cite{rawhani2024efficient} over 20 domain shifts.
\begin{table}[htbp]
\caption{COMPREHENSIVE COMPARISON WITH CURRENT STATE-OF-THE-ART UDAPTER AND OUR PREVIOUS APPROACH (PREV.) (MACRO F1 SCORES IN \%)}
\centering
\begin{tabular}{|c|c|c|c|c|}
\hline
\textbf{SRC→TGT} & \textbf{UDAPTER} & \textbf{PREV.} & \textbf{UNION} & \textbf{UNIPELT} \\
\hline
F→G & 79.79 & 78.93 & 79.59 & 80.23 \\
F→S & 72.30 & 72.68 & 73.66 & 73.66 \\
F→TE & 71.59 & 76.58 & 77.64 & 76.47 \\
F→TR & 77.07 & 76.75 & 75.95 & 76.89 \\
G→F & 73.56 & 74.09 & 74.99 & 74.96 \\
G→S & 71.36 & 71.69 & 72.99 & 73.07 \\
G→TE & 71.99 & 73.97 & 75.52 & 73.99 \\
G→TR & 76.79 & 76.80 & 76.99 & 77.48 \\
S→F & 75.35 & 75.80 & 77.33 & 76.40 \\
S→G & 80.94 & 81.15 & 82.25 & 80.39 \\
S→TE & 73.38 & 76.02 & 77.45 & 77.39 \\
S→TR & 77.16 & 77.94 & 78.02 & 78.09 \\
TE→F & 73.89 & 74.06 & 76.56 & 75.97 \\
TE→S & 71.41 & 71.52 & 72.43 & 73.48 \\
TE→G & 79.78 & 79.31 & 79.78 & 80.39 \\
TE→TR & 75.95 & 73.78 & 75.80 & 75.61 \\
TR→F & 73.13 & 72.43 & 73.04 & 74.27 \\
TR→S & 71.28 & 71.03 & 72.14 & 71.89 \\
TR→G & 81.55 & 80.32 & 80.27 & 80.38 \\
TR→TE & 71.42 & 74.11 & 75.35 & 74.61 \\
\hline
\textbf{Average} & \textbf{74.98} & \textbf{75.44} & \textbf{76.39} & \textbf{76.28} \\
\hline
\end{tabular}
\label{tab:udapter_comparison}
 
\end{table}

Our approach with the custom union achieves the highest average performance across domain shifts, with an average F1 score of 76.39\%. This represents an improvement of 1.41 percentage points over UDApter and 0.95 percentage points over our previous sequential approach. More specifically, UNION outperforms UDApter in 14 out of 20 domain shifts, while UNIPELT outperforms UDApter in 16 out of 20 domain shifts. When the two proposed mixed-PEFT variants are considered together, at least one of UNION or UNIPELT outperforms UDApter in 17 out of 20 domain shifts. These results indicate systematic advantages, demonstrate the robustness and effectiveness of the proposed strategy, and support its generalizability across different domain adaptation settings. Overall, the findings establish a new state-of-the-art result in parameter-efficient unsupervised domain adaptation under this evaluation setting.

\subsection{Effectiveness of Mixed-Objective Strategy with PEFT Combinations}

Our approach demonstrates the applicability of mixed-objective training in different well-designed PEFT combinations, not just ours. We evaluate two variants: our Custom Union (UNION) and UniPELT (UNIPELT), both trained using the mixed-objective methodology.

As shown in Table~\ref{tab:udapter_comparison}, our Custom Union achieves a marginal advantage of 0.11 percentage points (76.39\% vs 76.28\%) over UniPELT, indicating a small performance advantage for the custom union \cite{rawhani2024efficient}. However, the competitive performance of both mixed-PEFT approaches with mixed-objective training provides compelling evidence of the effectiveness of the mixed-objective approach for domain adaptation when properly integrated with well-designed PEFT combinations.

\subsection{Single PEFT Method Limitations in Mixed-Objective Training}

To validate the necessity of PEFT combinations in mixed-objective training, we conducted some ablation experiments on two domain shifts comparing single PEFT methods using invertible adapter (inv) alone against our combination approaches. The results are reported in Table~\ref{tab:single_vs_combination}.

\begin{table}[htbp]
\caption{SINGLE PEFT VS. PEFT COMBINATIONS IN MIXED-OBJECTIVE TRAINING}
\centering
\begin{tabular}{|c|c|c|c|}
\hline
\textbf{DOMAIN SHIFT} & \textbf{SINGLE INV.} & \textbf{UNION} & \textbf{UNIPELT} \\
\hline
G→S & 72.26 & 72.99 & 73.07 \\
S→TR & 77.08 & 78.02 & 78.09 \\
\hline
\end{tabular}
\label{tab:single_vs_combination}
 
\end{table}

The results suggest that: (1) Single PEFT methods are insufficient as they cannot fully exploit the potential of mixed-objective training, proving our hypothesis of the critical contribution of unifying complementary PEFT methods, the ablation studies in our study \cite{rawhani2024efficient} further validates the outcome. (2) Synergistic effects are crucial: The interaction between different PEFT methods provides benefits that neither can achieve independently. (3) Architectural diversity is necessary: Combinations targeting multiple transformer components enable more comprehensive domain adaptation.

\subsection{Comprehensive Comparison with Traditional Fully-Tuned UDA Methods}

From Table \ref{tab:traditional_comparison}, we observe that our approach not only matches but exceeds the performance of traditional fully-tuned methods on overall average. We achieve improvements of 1.26 percentage points over DANN and 0.86 percentage points over DSN while using only 7\% of the model parameters. Across the 20 domain shifts, at least one of UNION or UNIPELT outperforms DANN in 17 cases and DSN in 16 cases. Moreover, in 14 cases, at least one of the proposed variants exceeds both traditional baselines. This represents an advancement in parameter-efficient unsupervised domain adaptation, demonstrating that carefully designed PEFT combinations with mixed-objective training can outperform computationally expensive fully-tuned approaches.

\begin{table}[htbp]
\caption{COMPARISON WITH TRADITIONAL UDA METHODS (DANN, DSN) AND UPPER BOUND (MACRO F1 SCORES IN \%)}
\centering
\begin{tabular}{|c|c|c|c|c|c|}
\hline
\textbf{SRC→TGT} & \textbf{UB} & \textbf{DANN} & \textbf{DSN} & \textbf{UNION} & \textbf{UNIPELT} \\
\hline
F→G & 82.19 & 79.17 & 79.79 & 79.59 & 80.23 \\
F→S & 74.09 & 73.68 & 72.36 & 73.66 & 73.66 \\
F→TE & 78.41 & 73.72 & 75.07 & 77.64 & 76.47 \\
F→TR & 81.81 & 76.99 & 76.82 & 75.95 & 76.89 \\
G→F & 78.59 & 73.41 & 72.62 & 74.99 & 74.96 \\
G→S & 74.09 & 72.51 & 71.93 & 72.99 & 73.07 \\
G→TE & 78.41 & 71.52 & 72.90 & 75.52 & 73.99 \\
G→TR & 81.81 & 77.42 & 77.80 & 76.99 & 77.48 \\
S→F & 78.59 & 75.91 & 76.62 & 77.33 & 76.40 \\
S→G & 82.19 & 80.91 & 81.27 & 82.25 & 80.39 \\
S→TE & 78.41 & 74.32 & 74.27 & 77.45 & 77.39 \\
S→TR & 81.81 & 76.81 & 78.17 & 78.02 & 78.09 \\
TE→F & 78.59 & 75.07 & 75.17 & 76.56 & 75.97 \\
TE→G & 82.19 & 78.57 & 79.24 & 79.78 & 80.39 \\
TE→S & 74.09 & 71.65 & 72.16 & 72.43 & 73.48 \\
TE→TR & 81.81 & 75.72 & 77.29 & 75.80 & 75.61 \\
TR→F & 78.59 & 73.22 & 72.44 & 73.04 & 74.27 \\
TR→G & 82.19 & 80.91 & 81.67 & 80.27 & 80.38 \\
TR→S & 74.09 & 70.76 & 70.97 & 72.14 & 71.89 \\
TR→TE & 78.41 & 70.41 & 71.98 & 75.35 & 74.61 \\
\hline
\textbf{Average} & \textbf{79.02} & \textbf{75.13} & \textbf{75.53} & \textbf{76.39} & \textbf{76.28} \\
\hline
\end{tabular}
\label{tab:traditional_comparison}

\end{table}

\textbf{Upper Bound Analysis:} Our approaches achieved 96.7\% (UNION) and 96.5\% (UNIPELT) of the upper bound performance, representing exceptional efficiency given the parameter constraints and the unavailability of labeled target data.

\subsection{Standard Deviation Analysis and Robustness}

As shown in Table~\ref{tab:std_analysis}, our methods demonstrate comparable variability to state-of-the-art approaches while achieving superior performance. \textbf{Comparable Stability:} UNION (0.45) and UNIPELT (0.43) show standard deviations similar to those of the established methods (0.35-0.48), indicating that our performance improvements do not come at the cost of increased instability.

\begin{table}[htbp]
\caption{STANDARD DEVIATION COMPARISON ACROSS METHODS}
\centering
\begin{tabular}{|c|c|c|c|}
\hline
\textbf{METHOD} & \textbf{AVG. STD} & \textbf{PERFORMANCE} & \textbf{S/P. RATIO} \\
\hline
DANN & 0.48 & 75.13 & 0.0064 \\
DSN & 0.35 & 75.53 & 0.0046 \\
UDApter & 0.35 & 74.98 & 0.0047 \\
UNION (Ours) & 0.45 & 76.39 & 0.0059 \\
UNIPELT (Ours) & 0.43 & 76.28 & 0.0056 \\
\hline
\end{tabular}
\label{tab:std_analysis}
 
\end{table}

\subsection{Parameter Efficiency and Computational Analysis}

From Table~\ref{tab:parameter_efficiency}, full fine-tuning achieves the highest F1 score but updates all model parameters and relies on labeled target data. In contrast, our custom PEFT union achieves 76.39\% F1 while updating only 7\% of the parameters and using no labeled target data. It also outperforms the fully tuned DANN and DSN baselines, and improves over UDApter by 1.41 percentage points with only a negligible increase in trainable-parameter ratio.

\begin{table}[htbp]
\caption{PARAMETER EFFICIENCY COMPARISON}
\centering
\begin{tabular}{|c|c|c|c|c|}
\hline
\textbf{METHOD} & \textbf{TOTAL} & \textbf{TRAINABLE} & \textbf{\%} & \textbf{F1} \\
\hline
Full FT & 110M & 110M & 100.0\% & 79.02 \\
DANN (Full) & 110M & 110M & 100.0\% & 75.13 \\
DSN (Full) & 110M & 110M & 100.0\% & 75.53 \\
UDApter & 110M & 7.1M & 6.5\% & 74.98 \\
Our Union  & 110M & 7.68M & 7.0\% & 76.39 \\
UniPELT  & 110M & 11.08M & 10.2\% & 76.28 \\
\hline
\end{tabular}
\label{tab:parameter_efficiency}
 
\end{table}

\section{Conclusion}

This paper presents a novel mixed-objective training approach for unsupervised domain adaptation that achieves significant improvements over existing methods while maintaining exceptional parameter efficiency. Our approach combines invertible adapters with Low-Rank Adaptation and employs simultaneous optimization of classification and masked language modeling objectives.

The experimental results demonstrate substantial improvements of 1.41 percentage points over the current parameter-efficient state-of-the-art UDApter, and outperform results of traditional fully tuned methods DANN and DSN while using only 7\% of the model parameters. The key contributions include the introduction of a well-designed mixed-PEFT model architecture utilizing a mixed-objective training strategy specifically for domain adaptation, which demonstrates improved results and establishes a new state-of-the-art result for parameter-efficient unsupervised domain adaptation.

Future research directions include extending our approach to other NLP tasks beyond natural language inference, exploring the application to larger language models, and investigating adaptive methods for automatically selecting optimal PEFT combinations based on domain characteristics.

\end{document}